\documentclass[10pt,twocolumn,letterpaper]{article}
\pdfoutput = 1
\usepackage{iccp}
\usepackage{times}
\usepackage{graphicx}
\usepackage{cite}
\usepackage{color}
\usepackage{epsfig}
\usepackage{graphicx}
\usepackage{amsmath}
\usepackage{comment}
\usepackage{mathrsfs}
\usepackage{amsthm}
\usepackage{amssymb}
\usepackage{amsfonts}
\usepackage{fullpage}
\usepackage{wrapfig}
\usepackage[boxed]{algorithm2e}
\usepackage[normalem]{ulem}
\usepackage{enumitem}
\usepackage[hidelinks]{hyperref}
\usepackage{url}

\usepackage{subfigure}

\newcommand{\bfx}{{\bf x}}

\newcommand{\bfy}{{\bf y}}

\newcommand{\bfe}{{\bf e}}
\newcommand{\bfX}{{\bf X}}

\newcommand{\reals}{{\mathbb R}}

\iccpfinalcopy % *** Uncomment this line for the final submission

\def\iccpPaperID{0006} % *** Enter the ICCP Paper ID here

\ificcpfinal\fi

\begin{document}

\title{\textit{LiSens} --- A Scalable Architecture for Video Compressive Sensing} % Replace with your title
\author{Jian Wang,$^\dagger$ Mohit Gupta,$^\ddag$ and Aswin~C.~Sankaranarayanan$^\dagger$\\
$^\dagger$ ECE Department, Carnegie Mellon University, Pittsburgh, PA\\
$^\ddag$ CS Department, Columbia University, New York, NY
}

\maketitle
%
% \twocolumn[{
%    \begin{@twocolumnfalse}
%      \maketitle
%\includegraphics[width=\textwidth]{figures/Image_on_sensor.pdf}
%    \end{@twocolumnfalse}
%   }]
% 
%
%

\thispagestyle{empty}

\begin{abstract}
The measurement rate of cameras that take spatially multiplexed
measurements by using spatial light modulators (SLM) is often
limited by the switching speed of the SLMs. This is especially true
for single-pixel cameras where the photodetector operates at a rate
that is many orders-of-magnitude greater than the SLM. We study the
factors that determine the measurement rate for such spatial
multiplexing cameras (SMC) and show that increasing the number of
pixels in the device improves the measurement rate, but there is an
optimum number of pixels (typically, few thousands) beyond which the
measurement rate does not increase. This motivates the design of
\textit{LiSens}, a novel imaging architecture, that replaces the
photodetector in the single-pixel camera with a 1D linear array or a
line-sensor. We illustrate the optical architecture underlying
LiSens, build a prototype, and demonstrate results of a range of indoor
and outdoor scenes. LiSens delivers on the promise of SMCs: imaging
at a megapixel resolution, at video rate, using an inexpensive
low-resolution sensor.

\end{abstract}

\section{Introduction} \label{sec:intro}

Many computational cameras achieve novel capabilities by using
spatial light modulators (SLMs). Examples include light field cameras
\cite{marwah2013compressive,tambe2013towards}, lensless cameras
\cite{huang2013lensless}, and compressive cameras
\cite{duarte2008single,hitomi2011video,reddy2011p2c2,lin2014dual}.
The measurement rate of these cameras is often limited by the
switching speed of the SLM used. This is especially true for
the compressive single-pixel camera (SPC) \cite{duarte2008single} where
a photodetector is super-resolved by a digital micromirror device
(DMD). The SPC inherits both the spatial resolution of the DMD as
well as its operating rate. On one hand, this allows capturing
images at a high resolution in spite of the sensor being a single
pixel. This is especially beneficial while imaging in non-visible
wavebands, where the combination of a single photodetector coupled
with a high-resolution DMD often provides an inexpensive alternative
to full-frame sensors.\footnote{Megapixel sensors in short-wave
infrared, typically constructed using InGaAs, cost more than USD
100k.} On the other hand, the SPC also inherits the operating rate
of the DMD which, for commercially-available units, is in tens of
kHz. At this measurement rate, acquiring images and videos at the
spatial resolution of the DMD is feasible only at a very low
temporal resolution.

\begin{figure}[!ttt]
\center
\includegraphics[width=0.475\textwidth]{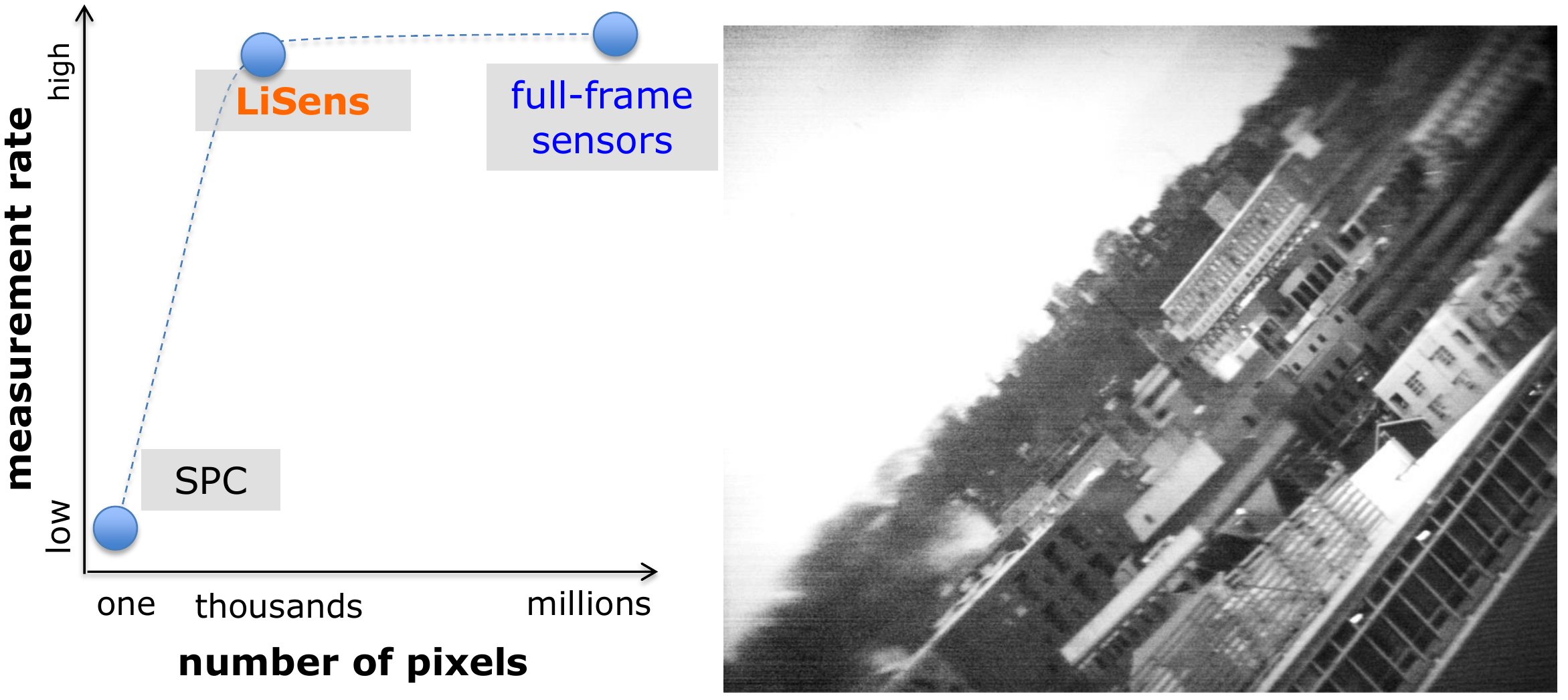}
\caption{\textbf{Motivation.} Varying the resolution of the sensor in a multiplexing camera provides interesting tradeoffs. A single-pixel camera is inexpensive but has limited measurement rate.
Full-frame sensors  deliver  high measurement rates but often at a steep price especially in non-visible wavebands.
The proposed camera, LiSens, uses a 1024 pixel line-sensor and achieves a measurement rate in MHz. (right) A $1024 \times 768$ pixel image of an outdoor scene using our prototype.}
\label{fig:motivation}
\end{figure}

%The achievable spatial and temporal resolutions for a conventional camera is determined by its measurement rate.
%%
%Specifically, the product of the spatial resolution,  in pixels, and the temporal resolution, in frames per second, cannot be greater than the measurement rate, in pixels per second.
%%
%The calculations are similar for the SPC except that its measurement rate is artificially amplified by the compression due to the use of signal priors.
%%
%If an SPC has a measurement rate of $M$ measurements per second and is capable of recovering images/videos at a compression of $C \ge 1$, then the product of the spatial and temporal resolutions can be no greater than $MC$.
%%
%State-of-the-art image and video compressive sensing (CS) algorithms seldom achieve compressions beyond 10$\times$ and 100$\times$, respectively.
%%
%To sense at a moderate spatial resolution, say a megapixel, and at video rate, say 30 frames per second, requires measurement rates in 10s of megapixels per second; this is many orders of magnitude beyond what is currently possible.  \note{To mohit: I think this entire para can be removed with no loss in continuity. or maybe moved to section 3.}

It is instructional to compare the two widely differing models of
sensing (see Figure \ref{fig:motivation}) in terms of their cost, determined by the number of pixels in the camera, and
measurement rate, defined as the number of measurements that the device obtains in unit-time.
Full-frame 2D sensors that rely on
Nyquist sampling are capable of achieving measurement rates in tens
of MHz, but are prohibitively expensive in many non-visible
wavebands. The SPC, which uses multiplexing as opposed to sampling,
has low measurement rates, but is inexpensive for exotic wavebands.
However, these imaging models are only the two extremes of a
continuous camera design space with varying number of pixels. Our
main observation is that the design that achieves the maximum
measurement rate at low cost lies between the two extremes.

Based on this observation, we propose a novel architecture for video
compressive sensing (CS) that is capable of delivering measurement
rates in MHz. The proposed camera, which we call
\textit{LIne-SENSor-based compressive camera (LiSens)}, uses a 1D
array of pixels to observe a scene via a DMD. Each pixel on the
sensor is optically mapped to a row of micromirrors on the DMD and
hence, the camera obtains a coded line-integrals of the image formed
on the DMD. The high measurement rate of LiSens enables CS of images
and videos at  high spatial resolutions and reasonable
frame-rates. We provide the optical schematic for building such a
camera and illustrate its novel capabilities with a lab prototype.

\section{Prior work} \label{sec:prior}

\paragraph{Multiplexed imaging.}
This paper relies heavily on the concept of optical multiplexing.
Suppose that we seek to sense a signal $\bfx \in \reals^N$.
A conventional camera samples this signal directly and has an image formation model given by
\[ \bfy = \bfx + \bfe,\]
where $\bfe$ is the measurement noise.
%
%Given $\bfy$, the estimate of $\bfx$, $\widehat{\bfx} = \bfy$, has an error of $\| \bfe \|$.
%
Multiplexing  expands the scope of the camera by allowing it to obtain any arbitrary linear transformation of the signal:
\[ \bfy = A \bfx + \bfe. \]
Assuming that $A \in \reals^{N \times N}$ is invertible,  the signal can be estimated as  $\widehat{\bfx} = A^{-1} \bfy$.
It is  well known that, when the entries of $A = [a_{ij}]$ are restricted to $| a_{ij}| \le 1$, using a Hadamard measurement matrix is optimal provided the noise is signal independent \cite{harwit1979hadamard}.
Note that in the absence of any priors, we need at least $N$ measurements  to recover an $N$-dimensional signal.
%the estimate $\widehat{\bfx} = A^{-1} \bfy$ incurs an error equal to $\| \bfx - \widehat{\bfx} \| = \| A^{-1} \bfe \| \le \| A^{-1} \| \| \bfe \|$.
%
%Passive sensing requires that the measurement matrix $A = [a_{ij}]$ have entries with magnitude less than or equal to 1, i.e,
%$ | a_{ij} | \le 1. $

%%
%In particular, when $\bfy = H \bfx + \bfe$, where $H \in \reals^{N \times N}$ is the $N$-dim.\ Hadamard matrix, then the estimate of $\bfx$, $\widehat{\bfx} = H^T \bfy/N$ has an error equal to $\| \bfx - \widehat{\bfx} \| = \| \bfe \|/\sqrt{N}$.
%%
%This reduction in the error from $\| \bfe \|$ to $\| \bfe \|/\sqrt{N}$ is termed the \textit{multiplexing gain} and provides substantial improvement in the quality of the reconstructed signal especially for large values of $N$.
%%
%The price we pay is often in the form of complex imaging architecture required to optically implement the multiplexing scheme.

%Hadamard multiplexing has found widespread use in many applications including spectroscopy \cite{nelson1970hadamard,marshall1975fourier,decker1969experimental}, infrared imaging \cite{davies1975spatially}, and light transport acquisition \cite{schechner2007multiplexing}.
%%
%The multiplexing gain can be used to obtain improved reconstruction quality or can be traded off to obtain faster acquisition without loss in quality.
%%
%Further, for architectures like the SPC,  even with the  gain provided by Hadamard multiplexing, the time to recover a $N$-dimensional scales as $O(\sqrt{N})$ --- which leads to very poor temporal resolutions for large values of $N$.

\vspace{-3mm}
\paragraph{Compressive sensing (CS).}
CS aims to sense a signal $\bfx \in \reals^N$ from an under-determined linear system, i.e., measurement  of the form \[ \bfy = \Phi \bfx + \bfe, \]
where $\Phi \in \reals^{M \times N}$ with $M < N$.
For an arbitrary signal in $\reals^N$ this is impossible since the map $\Phi: \reals^N \mapsto \reals^M$ is many-to-one and non-invertible.
CS handles this by restricting the signal  $\bfx$ to belong to a distinguished class; for example, sparse signals in a transform basis.
The main results of CS state that when the measurement matrix $\Phi$ has a special structure and $\bfx$ is $K$-sparse in a transform basis, then we can robustly recover $\bfx$, provided $M = O(K \log (N/K))$ \cite{baraniuk2007compressive}.
%
%These basic results have been extended beyond sparsity to include signals that have sparse gradients \cite{chambolle2004algorithm,osher2005iterative}, low rank matrices \cite{candes2009exactr}, and even low-dimensional manifolds \cite{shah2011iterative}.
%

\vspace{-3mm}
\paragraph{Video models for CS.}
There have been many models proposed to handle  CS of videos.
Wakin et al.\ \cite{wakin2006compressive} proposed the use of 3D wavelets as a sparsifying transform for videos.
This model is further refined in Park and Wakin \cite{park2009multiscale} where a lifting scheme is used to tune a wavelet to better approximate the temporal variations.
Dictionary-based models were used in Hitomi et al.\ \cite{hitomi2011video} for temporal resolution of videos; it is shown that a highly overcomplete dictionary provides high quality reconstructions.
Inspired by the use of motion-flow models in video compression, Reddy et al.\ \cite{reddy2011p2c2} and subsequently, Sankaranarayanan et al.\ \cite{sankaranarayanan2012cs} employed an iterative strategy where optical flow derived from an initial estimate is used to further constrain the video recovery problem.
This provides a significant improvement in reconstruction quality, albeit at a high computational cost.
Gaussian mixture models were used in \cite{yang2014video} for video CS; a hallmark of this approach is that the  mixture model is learned directly from the compressive measurements providing a model that is  tailored to the specifics of the scene being reconstructed.

%In addition to novel signal models and recovery algorithms, the  ideas in CS have also inspired a suite of novel imaging architectures.
%
%We discuss this next.

\vspace{-3mm}
\paragraph{Compressive imaging hardware prototypes.}
The original prototype for the SPC used a DMD as the programmable SLM \cite{duarte2008single}.
In \cite{huang2013lensless}, a variant of the SPC is proposed where an LCD panel is used in place of the DMD; the use of a transmissive light modulator enables a lensless architecture.
Sen and Darabi \cite{sen2009compressive} use a camera-projector system to construct an SPC exploiting a  concept called dual photography \cite{sen2005dual}; the hallmark of this system is its use of active illumination.

There have been many multi-pixel extensions to the SPC.
The simplest approach \cite{kelly2014decreasing} is to map the DMD to a low-resolution sensor array, as opposed to a single photodetector, such that each pixel on the sensor observes a non-overlapping ``patch'' or a block of micromirrors on the DMD.\footnote{The interested reader is referred to earlier work by Nayar et al.\ \cite{nayar2004programmable} where a scene is observed by a sensor via a DMD to enable programmable imaging. However, this device was not used for spatial multiplexing or compressive sensing.}
SMCs based on this design have been proposed for sensing in mid-wave infrared camera \cite{mahalanobis2014recent} and short-wave infrared \cite{chen2015fpacs}.
Measurement matrix designs for such block-based compressive imaging architectures are presented in Kerviche et al.\ \cite{kerviche2014information} and Ke and Lam \cite{ke2012object}.

While we focus on spatial multiplexing, many architectures have been proposed for multiplexing other attributes of a scene including temporal \cite{gao2014single,hitomi2011video,reddy2011p2c2,llull2013compressive,holloway2012flutter}, angular \cite{marwah2013compressive,tambe2013towards}, and spectral \cite{wagadarikar2008single,lin2014dual}.
All of these architectures use a high-resolution sensor and  sacrifice  this spatial resolution partially to obtain higher resolution in time, spectrum and/or angle.
In contrast, the imaging architecture proposed in this paper takes a device with a high temporal resolution and sacrifices the temporal resolution in part to obtain higher spatial resolution.

%\textit{Temporal multiplexing.} The idea of temporal multiplexing camera is to take a device with low temporal resolution or frames per sec and obtain a high-speed camera from it.
%%
%Veeraraghavan et al.\ \cite{} use a flutter shutter camera \cite{} or a camera with coded global shutter to convert a 30fps device to a 2000 fps device provided the scene exhibited periodic motion.
%%
%These results were extended by Halloway et al.\ \cite{} for non-periodic scenes but only for a modest temporal super-resolution.
%%
%Reddy et al.\ \cite{} and Hitomi et al.\ \cite{} propose variants of a per-pixel coded shutter, where the temporal profile at each pixel is independently modulated.
%%
%These are in many ways the temporal dual to the SPC.
%
%\textit{Spectral multiplexing.}

\section{Measurement rate of a camera} \label{sec:tradeoffs}

The measurement rate of a camera is given by the product of its spatial resolution, in pixels, and the temporal resolution, in frames per second.
A spatial multiplexing camera (SMC) captures one or more coded linear measurements of a scene via a spatial light modulator (e.g., DMD).
Multiple measurements are taken sequentially by changing the code displayed on the DMD.
Thus, the measurement rate of an SMC is limited by the resolution and the frame-rate of the sensor, as well as the resolution and switching speed of the DMD.\footnote{Compressive cameras can further increase the spatial or temporal resolution by using scene priors.}
The operating rate of the DMD is determined by the rate at which the micromirrors can be switched from one code into another.
Denoting this rate as $R_{DMD}$ Hz, we note that the frame rate of the SMC cannot be greater than $R_{DMD}$ since the specific linear mapping from the scene to sensor is determined by the micromirror code.
The frame rate of a sensor is also limited by its readout speed --- typically, determined by the operating rate of the analog-to-digital converter (ADC) in the readout circuit.
Suppose that we have an ADC that can perform readout at  a rate of $R_{ADC}$ Hz.
Then, given a frame with $F$ pixels, the ADC limits the sensor to $R_{ADC}/F$ frames per second. Hence, the measurement rate of the
SMC is given as
\[ F \times \min \left( R_{DMD}, \frac{R_{ADC}}{F} \right)  \quad \textrm{meas.\ per sec.} \]

\begin{figure}[!ttt]
\center
\includegraphics[width=0.425\textwidth]{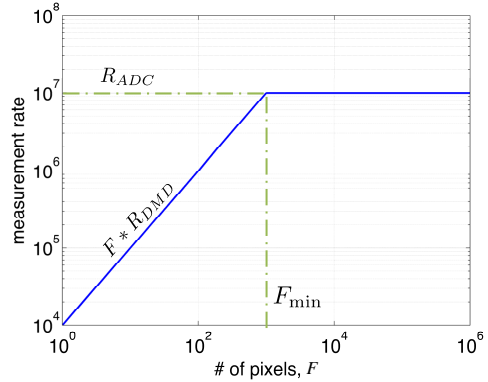}
\caption{\textbf{Measurement rate and sensor resolution.} There are two regimes of operations: for $F < F_{\min}$, the bottleneck is the DMD and for $F > F_{\min}$, the bottleneck is the sensor readout. At $F = F_{\min}$ lies a design that obtains the highest measurement rate with the  least sensor resolution.}
\label{fig:measrate}
\end{figure}

Operating speeds of  commercially-available DMDs are 2-3 orders of magnitude lower than that of ADCs in sensors.
As a consequence, for small values of $F$, the measurement rate of an SMC is dominated by the operating speed of the DMD (see Figure \ref{fig:measrate}).
Once $F \ge R_{ADC}/R_{DMD}$, then the bottleneck shifts from the DMD to the ADC. Further, the smallest sensor resolution (in terms of
number of pixels) for which the measurement rate is maximized is \[
\textrm{(minimum number of  pixels)}\quad  F_{\min} = R_{ADC}/R_{DMD}.\]

\textit{In essence, at $F = F_{\min}$ we can obtain the measurement
rate of a full-frame sensor but with a device with potentially a
fraction of the number of pixels.} This is invaluable for sensing in
many wavebands, for example short-wave infrared.

As a case study, consider an SMC with a DMD operating rate $R_{DMD} =
10^4$ Hz and an ADC with an operating rate $R_{ADC} = 10^7$ Hz.
Then, for a sensor with  $F_{\min} = 10^3$ pixels, we can obtain
$10^7$ measurements per second. An SPC, in comparison, would only
provide  $10^4$ measurements per second. This significant increase
in measurement rate motivates the multi-pixel SMC design that we
propose in the  next section.

\section{LiSens} \label{sec:linesense}

%
%We refer to this camera as \textit{\underline{LI}ne-\underline{Sens}or-based compressive camera (LiSens)}.

\begin{figure}[!ttt]
\center
\includegraphics[width=0.475\textwidth]{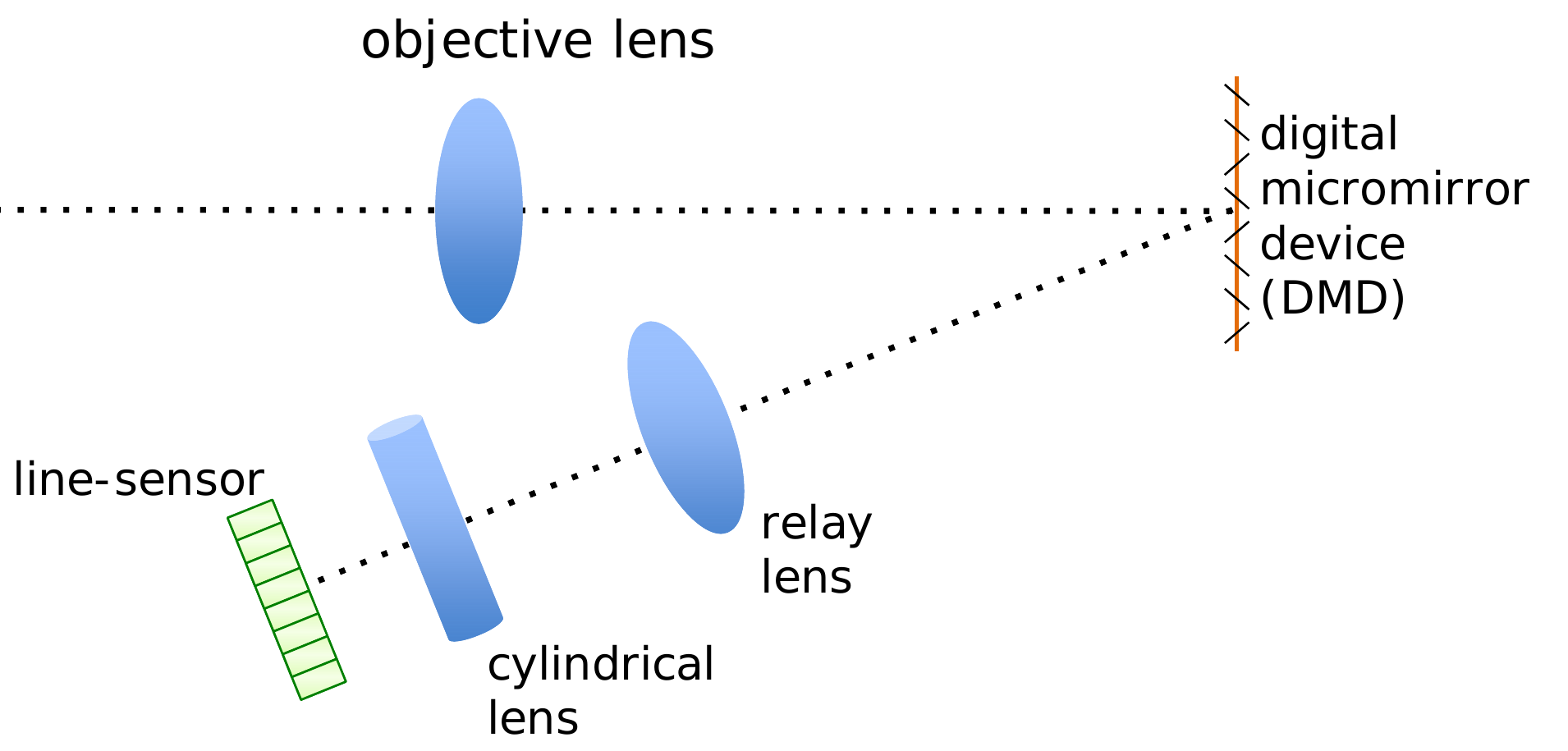}
\caption{{\bf Schematic of the LiSens imaging architecture.}}
\label{fig:schematic}
\end{figure}

\begin{figure}[!ttt]
\center
\includegraphics[width=0.475\textwidth]{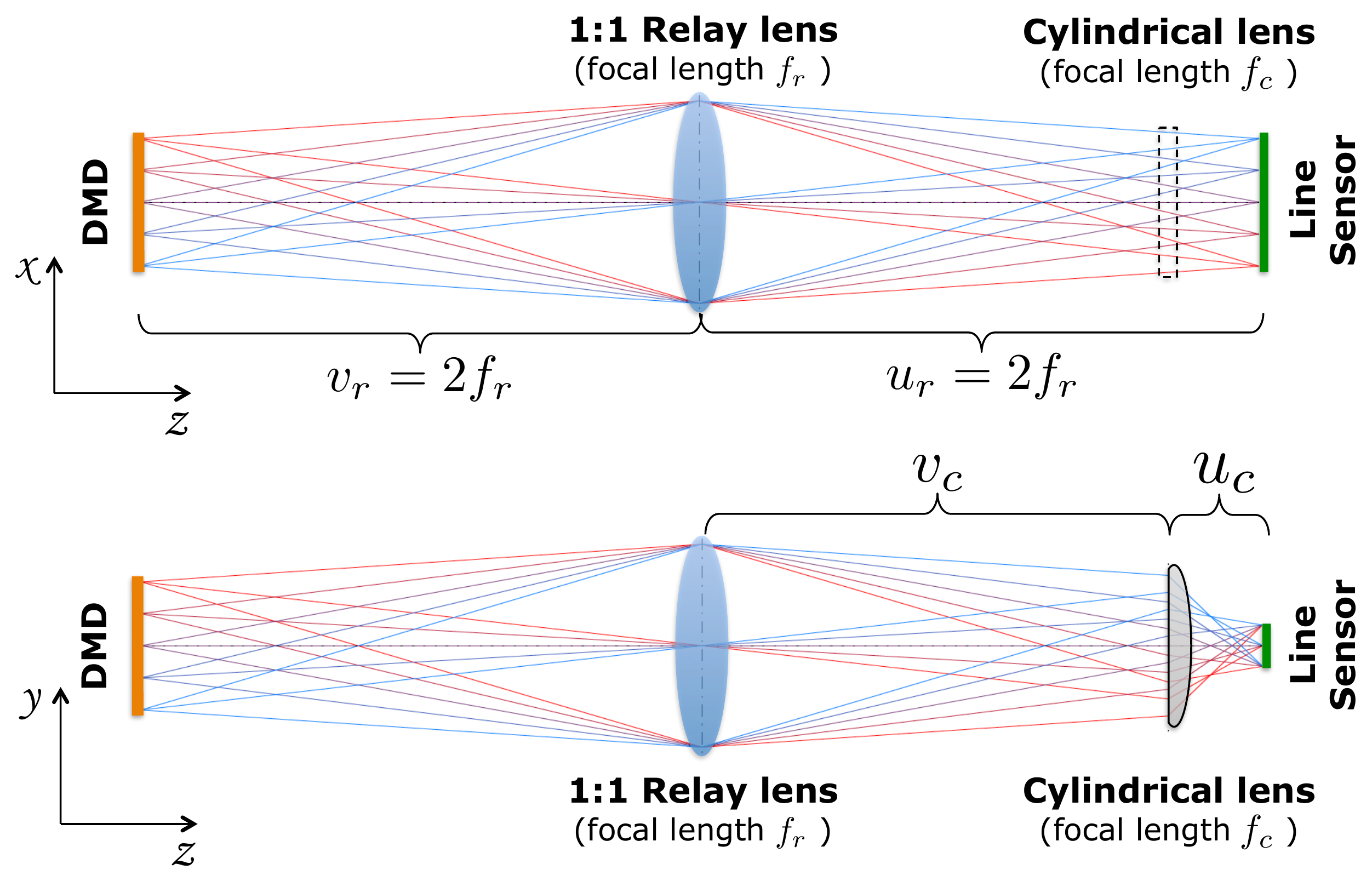}
\caption{{\bf The DMD-to-sensor mapping.} Shown are the ray diagrams for the mapping of the DMD to line-sensor along two orthogonal directions ---  (top) along and (bottom) orthogonal to the axis of cylindrical lens. The cylindrical lens  focuses the aperture plane of the relay lens onto the sensor plane. Hence, the sensor plane is focused on the DMD along one-axis (top) but defocused along the other (bottom).}
\label{fig:dmdsensor}
\end{figure}

The optical setup of LiSens is illustrated in Figure \ref{fig:schematic}.
%
%Light from a scene is focused onto a DMD.
%
The DMD splits the optical axis into two axes --- one each for $\pm11^\circ$ orientations of the micromirrors.
Along one of these axes, the 2D image of the scene formed on the DMD plane is mapped onto the 1D line-sensor using a relay lens and a cylindrical lens.
The image captured by the line-sensor can be represented as a 1D integral of the 2D image (along rows or columns) formed on the DMD plane.
This is achieved by aligning the axis of the cylindrical lens with that of the line-sensor and placing it so that it optically mirrors the aperture plane of the relay lens onto the sensor plane (see Figure \ref{fig:dmdsensor}).


\begin{figure}[!ttt]
\center
\includegraphics[width=0.475\textwidth]{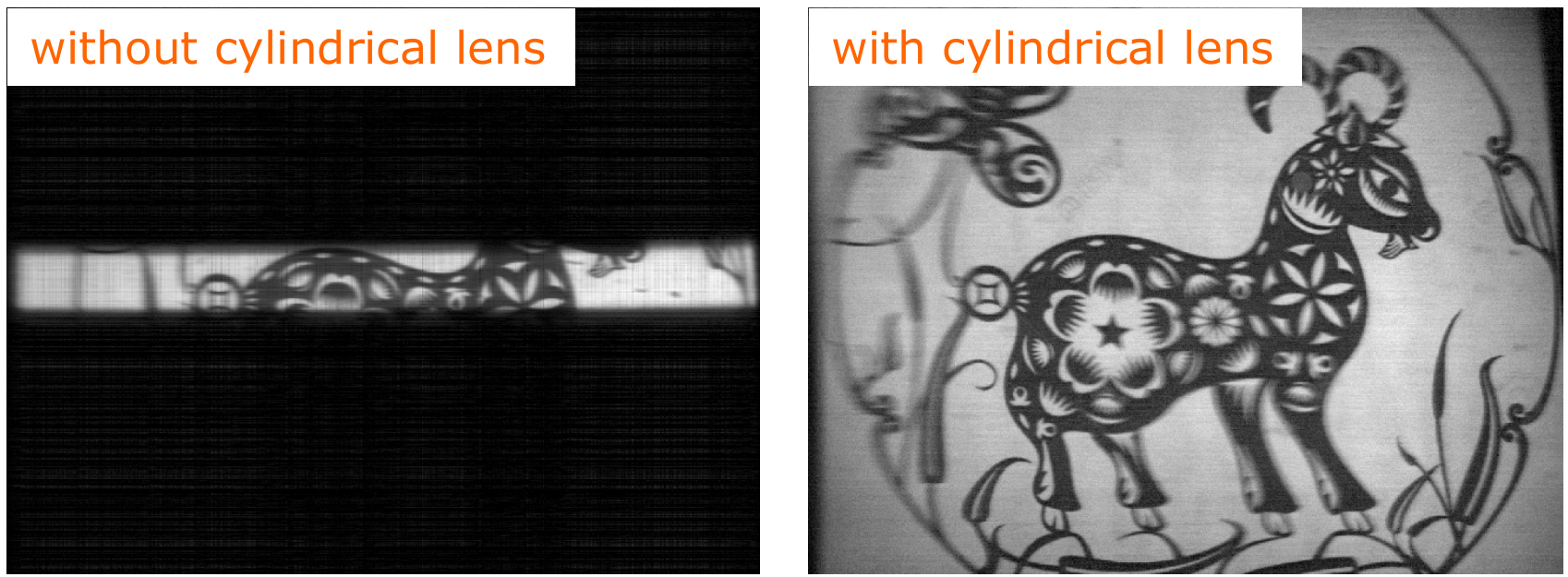}
\caption{{\bf Field-of-view enhancement  by the cylindrical lens.} Shown are images acquired with our prototype.}
\label{fig:cylinder}
\end{figure}

Suppose that the physical dimensions of the  DMD and the line-sensor  are $(w_D, h_D)$ and $(w_L, h_L)$, respectively, with $h_L \ll w_L$.
The relay lens is selected to produce a magnification of $w_L/w_D$ so that the DMD maps to the line-sensor along its width; this determines the ratio of the focal lengths used in the relay lens.
The exact value of the focal lengths is optimized to minimize vignetting and needs to be determined along with the objective lens.
If $w_D = w_L$,  we can use a 1:1 relay lens with a focal length $f_{r}$ and the distance between the relay lens and the line-sensor is $2 f_{r}$.
The parameters and the placement of the cylindrical lens are determined so that the aperture of the relay lens is magnified (or shrunk)  within the line-sensor.
This corresponds to a  magnification of $h_L/d_{r}$, where $d_r$ is the diameter of the relay lens.
Let $f_{c}$ be the focal length of the cylindrical lens, and suppose that it is placed at a distance of $u_c$ from the line-sensor and $v_c$ from the relay lens.
Then,
\[u_c + v_c  =2f_{r},\
\frac{1}{u_c} + \frac{1}{v_c} = \frac{1}{f_{c}},\
\frac{u_c}{v_c} = \frac{h_L}{d_{r}} \]
Hence,
\begin{equation}
f_{c} = 2f_r \frac{h_L}{d_r+h_L} \frac{d_r}{d_r+h_L} \approx 2 f_r \frac{h_L}{d_r}.
\label{eqn:focal_cyl}
\end{equation}
%and
%\[ u_c = 2f_c \frac{h_L}{d_r+h_L} \approx 4 f_r \left(\frac{h_L}{d_r}\right)^2. \]
%
In practice, we  observe that, for a marginal loss of light throughput, we could obtain flexibility in both the choice of the cylindrical lens (focal length) as well as its positioning.
Figure \ref{fig:cylinder} shows the improvement in the vertical field-of-view when the cylindrical lens is introduced.

\begin{figure}[!ttt]
\center
\includegraphics[width=0.475\textwidth]{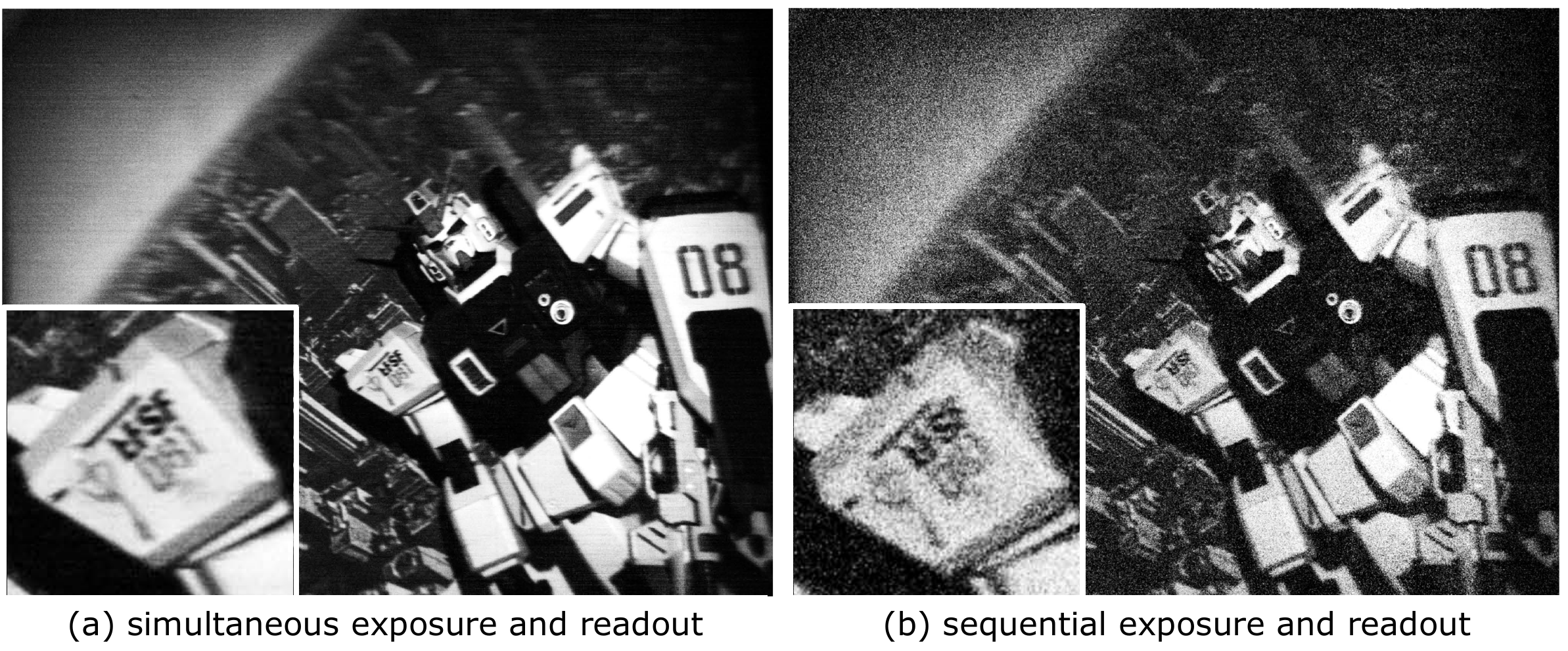}
\caption{{\bf Benefits of frame transfer.} The DMD was operated with $400 \mu s$ per pattern. The sensor had a readout of $350 \mu s$ per frame. (a) Simultaneous readout allows for a   $400 \mu s$ exposure. (b) Sequential readout cut the exposure time down to $50 \mu s$, resulting in a noisy image. Shown are images acquired with our prototype.}
\label{fig:frametransfer}
\end{figure}

\subsection{Practical advantages}\label{sec:benefit}
Before we proceed, it is worth pondering on alternative and potentially simpler designs for multi-pixel SMCs.
An intuitive multi-pixel extension of the SPC would be using a low-resolution 2D sensor array \cite{mahalanobis2014recent,chen2015fpacs}.
Having a 2D sensor array provides a simpler mapping from the DMD that can be achieved using just relay lenses.
Yet, there are important considerations that make the LiSens camera a powerful alternative to using 2D sensors.

%A line-sensor provides a suite of beneficial properties over a 2D sensor array.
%
Specifically, it is  simpler and inexpensive to obtain frame transfer\footnote{Frame transfer is a  technology used to enable simultaneous exposure and readout in a sensor. In a traditional camera, during the readout process, the sensor is not exposed; this reduces the duty cycle of the sensor and leads to inefficiencies in light collection. With frame transfer, there is a separate array of storage pixels beside the photosensitive array. After exposure, charge from the pixels are immediately transferred to the storage array allowing the sensor to be exposed again.} on a line sensor without requiring complex circuitry or loss of light due to reduced fill-factor.
Figure \ref{fig:frametransfer} demonstrates the benefits of frame transfer on our lab prototype.
The 1D profile of the sensor also provides the possibility of having a per-pixel ADC; this provides a dramatic increase in the readout rate of the sensor.
Finally, we benefit from the fact that line-sensor have long been  manufactured for spectroscopy which requires very precise, low-noise sensors with high dynamic range and broad spectral response; these properties make them highly desirable for our application.

\subsection{Imaging model}
Suppose the DMD has a resolution of $N \times N$ micromirrors, and the line-sensor has $N$ pixels such that each pixel maps to a row of micromirrors on the DMD.
At time instant $t$, let $\bfX_t \in \reals^{N \times N}$ be the scene image formed on the DMD, and
let $\Phi_t \in \reals^{N \times N}$ be the binary pattern displayed on the DMD.
Then, the measurement obtained at the line-sensor, $\bfy_t \in \reals^N$, is given as
\[ \bfy_t = (\bfX_t \circ \Phi_t ) {\bf 1}  + \bfe_t,\]
where $\circ$ denotes the entry-wise product, ${\bf 1}$ is the vector of ones and $\bfe_t$ is the measurement noise.

For simplicity, we use DMD patterns where every column is the same, i.e., patterns of the form
\[ \Phi_t = {\bf 1}{\phi_t}^T, \]
where $\phi_t \in \reals^N$ is a binary vector.
This alleviates the need for extensive calibration.
The sensor measurements are then given as
\[ \bfy_t = \bfX_t \phi_t + \bfe_t. \]
For the experiments in the paper, we use rows of a column-permuted Hadamard matrix for $\phi_t$.

\subsection{Recovery} \label{sec:time}

\paragraph{Sensing images.}
When the scene is static, i.e.,  $\bfX_t~=~\bfX$ for $t \in \{1, \ldots, T\}$, we can write the imaging model as
\[ {\bf Y} = [ \bfy_1, \ldots, \bfy_T] = \bfX \Phi + E,\]
where $\Phi = [\phi_1, \ldots, \phi_T]$ is the measurement matrix and $E = [\bfe_1, \ldots, \bfe_T]$ is the measurement noise.

If $T \ge N$ and $\Phi$ were well-conditioned, say for example, a Hadamard matrix, then we can obtain an estimate of the scene image as
\begin{equation}
\widehat{\bfX} = {\bf Y} \Phi^{\dagger}.
\label{eq:hada}
\end{equation}

In practice, it is unreasonable to expect a scene to be static over a large duration and hence, we can expect $T < N$.
To regularize the inverse problem, we enforce sparsity in the gradients of the recovered image using a minimum total-variation prior \cite{chambolle2004algorithm}.
This leads to the following optimization problem.
\begin{equation}
 \min_\bfX  TV(\bfX), \quad \textrm{s.t.} \quad \| {\bf Y} - \bfX \Phi \|_F \le \epsilon,
 \label{eq:TV}
 \end{equation}
where $TV(\bfX)$ is the total-variation norm that captures strength of the image gradients.
We used the isotropic TV-norm defined as
\[ TV(\bfX) = \sum_{i=1}^N \sum_{j=1}^N \sqrt{G_x^2(i, j) + G_y^2(i, j)}, \]
where $G_x, G_y \in \reals^{N \times N}$ are the spatial gradients of $\bfX$.

\vspace{-3mm}
\paragraph{Sensing videos.}
We enforce sparse spatio-temporal gradients when recovering time-varying scenes.
Specifically, suppose that we obtain measurements $\{\bfy_t, 1  \le t \le T \}$ for a time-varying scene $\{\bfX_t, 1 \le t \le T \}$.
It is often an overkill to recover an image for every frame of the sensor since the problem becomes computationally overwhelming.
Instead, we decide on a target frame-rate (say 10 frames per second) and group together successive measurements so as to obtain the desired frame rate.
Let $ \{ \widetilde{\bfX}_k, k = 1, \ldots, Q\}$ be the frames associated with the scene at the desired frame-rate.
Then, the imaging model reduces to
\[  t = k\frac{T}{Q},\quad [\bfy_{t},  \ldots, \bfy_{t+\frac{T}{Q}} ] = \widetilde{\bfX}_k  [\phi_{t}, \ldots, \phi_{t+\frac{T}{Q}} ] + E_k \]
Grouping together measurements and associating them to a frame of a video also reduces the number of parameters to be estimated.
Interested readers are referred to \cite{park2013multiscale} for a frequency domain analysis on judicious selection of frame-rate for the SPC.
Finally, similar to (\ref{eq:TV}), we solve for the video by minimizing a spatio-temporal TV-norm constrained by the measurements.

%\section{Design of measurement patterns} \label{sec:design}
%\input{design.tex}

\section{Hardware prototype} \label{sec:prototype}

%We built a LiSenS prototype for the visible waveband as a proof of concept.
%
Our proof-of-concept prototype was built using a Nikkor 50mm F/1.8 objective lens, a DLP7000 DMD, and a Hamamatsu S11156-2048-01 line-sensor.
The DMD had a spatial resolution of $1024 \times 768$ with a micromirror pitch of $13.8 \mu m$ and a maximum operating speed of $20$ kHz. 
The line sensor had 2048 pixels with a pixel size of $14 \mu m \times 1 mm$;  we were only able to use 1024 pixels due to our inability to find a cylindrical lens that was sufficiently long to span the entire line-sensor.
Due to the pixel pitch and micromirror pitch being nearly the same, we used a 1:1 relay lens and oriented the line sensor so that the line sensor was aligned to the width of the DMD.
Hence, each pixel on the line sensor summed up $768$ micromirrors on the DMD. 

Our line sensor provided access to simultaneous exposure and readout using frame transfer.
We operated the prototype with an exposure of approximately $500 \mu s$.
Due to buffer limitations in the readout circuit (a Hamamatsu C11165-01 driver), we were able to obtain $100$ frames at $500 \mu s$ per frame, followed by a cool-down time of $60 ms$.
As a consequence, the device provided  $900$ frames per second and hence, a measurement rate of $9\times10^5$ measurements per second.

\begin{figure}[!ttt]
\center
\includegraphics[width=0.475\textwidth]{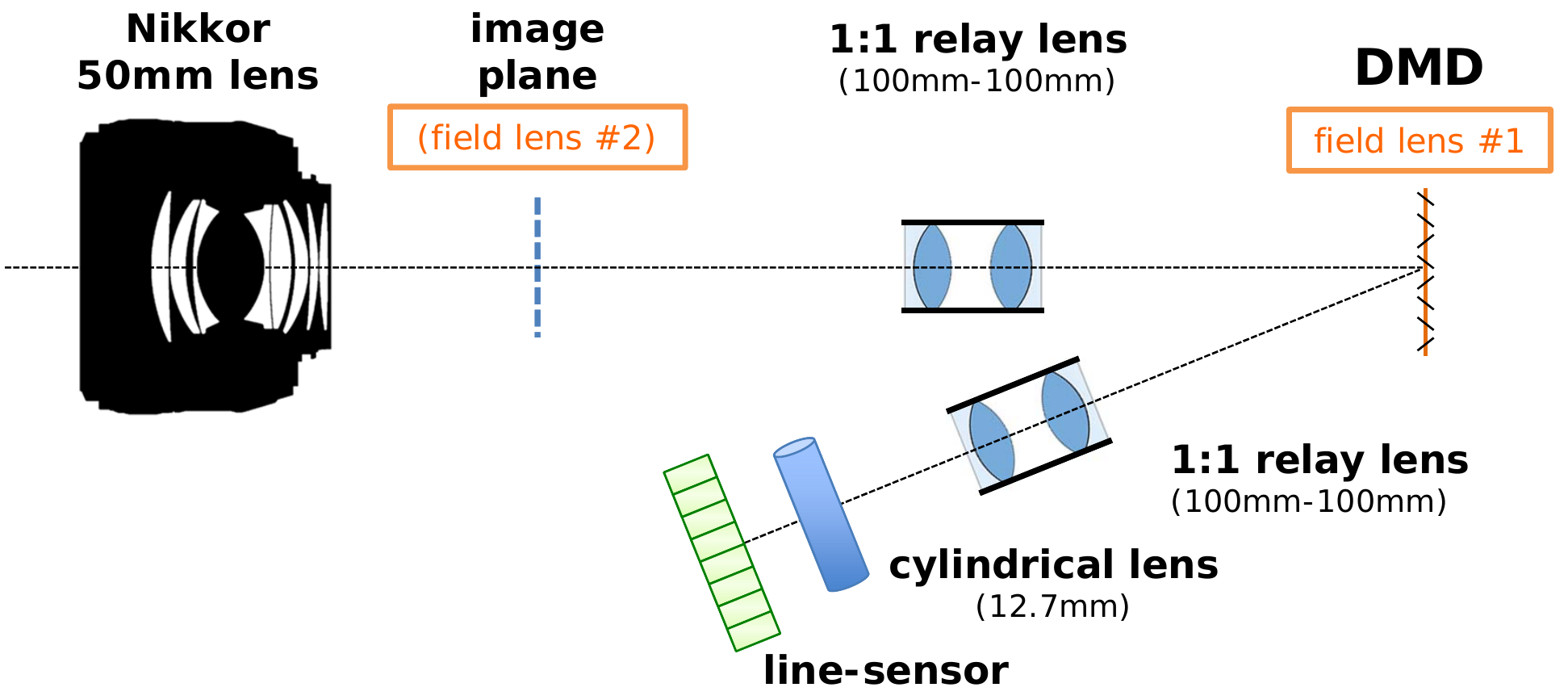} 
\includegraphics[width=0.475\textwidth]{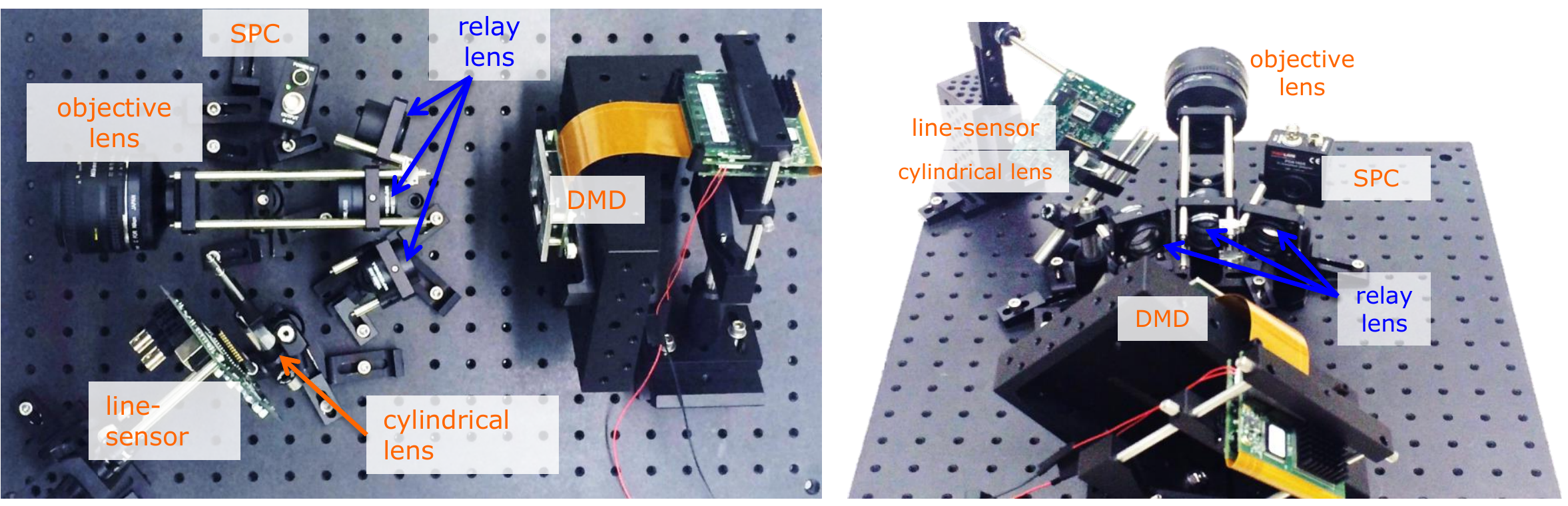}
\caption{{\bf Prototype of the LiSens camera.}}
\label{fig:prototype}
\end{figure}

For comparison and validation, we built an SPC on the off-axis of the DMD (see Figure \ref{fig:prototype}).
The measurement rate of the SPC was $20$ kHz, the speed of the DMD.

Figure \ref{fig:prototype} shows both the modified schematic of the camera as well as the optical layout.
The  subtle differences to the schematic shown in Figure \ref{fig:schematic} stem from a couple of practical constraints.
Recall that the DMD reflects the optical axis by $22^\circ$.
The flange distance of  objective lenses (F-mount and  C-mount) is insufficient to provide clearance for the cone of light reflected from the DMD.
To alleviate this, we optically mirror the image plane of the objective lens using a 1:1 relay lens, thereby providing ample space for the reflected cone of light.
We used a 100mm:100mm achromatic doublets for the relay lens, thereby providing an $f_r = 50mm$ and  a cylindrical lens with a focal length of $12.7mm$.\footnote{We used relay lenses with diameter $d_r = 25mm$. The value suggested in (\ref{eqn:focal_cyl}) is  $f_c = 4mm$. However, to avoid custom designed optics, we went with a larger focal length.}
% just shy of the $10mm$ requirement from (\ref{eqn:focal_cyl}).

%
%
%\begin{figure}[!ttt]
%\center
%\includegraphics[width=0.475\textwidth]{figures/FieldLenses.pdf}
%\caption{{\bf Field lenses and vignetting.} We look images recovered by an SPC looking at the DMD when field lenses are used. Shown are four different configurations corresponding to the presence/absence of  field lenses at the two locations highlighted in Figure \ref{fig:prototype}.}
%\label{fig:field}
%\end{figure}
%
%
%
%\begin{figure}[!ttt]
%\center
%\includegraphics[width=0.475\textwidth]{figures/Aperture_BvV.pdf}
%\caption{{\bf Vignetting versus F/\#.} Vignetting decreases as we increase the aperture of the main lens. However, the resulting image is also blurred.}
%\label{fig:aperture}
%\end{figure}
%

\vspace{-3mm}
\paragraph{Alignment.} 
Misalignment of the optical components, in addition to introducing blur,  results in a DMD column  mapping to multiple pixels on the line detector.
 In our prototype, we minimized both effects  by  manually adjusting each component while observing and quantifying sharpness of the images in real-time.
While this produces acceptable results, a calibration procedure is indeed required especially to resolve the image especially at the boundaries of the field of view.

\vspace{-3mm}
\paragraph{Vignetting.} The use of relay lenses to extend the optical axis causes significant vignetting.
To alleviate this, we introduced a  field lens at the DMD which dramatically reduces vignetting.
A second field lens at the image plane of the objective lens further reduces vignetting but with an  increase in spatial blur.
The results in this paper were produced with a single field lens at the DMD.

%\vspace{-3mm}

%
%To this end, for the results in the paper, we use a single field lens --- a plano-convex lens places on top of the DMD.
%
%Figure \ref{fig:aperture} looks at vignetting as the aperture is reduced.
%%
%Reducing the aperture does increase sharpness at the cost of reduced light throughput as well as increased vignetting.
%%
%For most of experiments in the paper we used an aperture of F/2.8 or F/1.8 on the main lens.
%%

%\vspace{-3mm}
%\paragraph{Calibration.}
%
%{\note{was ignored so far. have to explain why. I might look into this to see if we can deblur the images better ... if so, power!}}

\section{Experiments} \label{sec:results}

\begin{figure*}[!ttt]
\includegraphics[width=\textwidth]{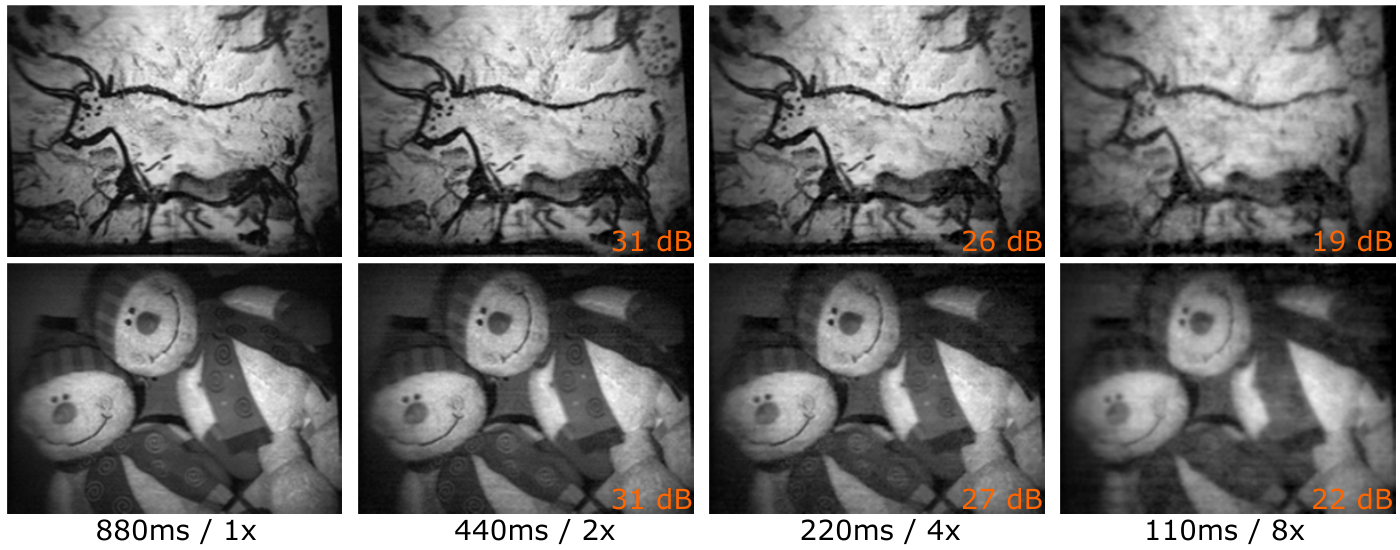}
\caption{\textbf{LiSens reconstructions for a few scenes under varying capture durations / under-sampling.} The reconstructions suffer little loss in quality until a capture time of $110 ms$; this corresponds to an under-sampling of $8\times$. Inset on each figure is the reconstruction SNR, in dB, using the $1\times$ reconstruction as the ground truth.}
\label{fig:gallery}
\end{figure*}

We use two metrics to characterize the operating scenarios for our experiments:

\vspace{-2mm}{\flushleft \textit{(i) The under-sampling}} given as the ratio of the dimensionality of the acquired image to the number of measurements, i.e., if we seek to acquire an $N_1 \times N_2$ image with $M$ measurements, then the under-sampling is $N_1N_2/M$. When the under-sampling  is $1$, the system is invertible and we use (\ref{eq:hada}) to recover the image. For values greater than $1$, we use the TV prior and solve (\ref{eq:TV}).

\vspace{-2mm}{\flushleft \textit{(ii) Temporal resolution / capture duration per frame.}} Recall, from Section \ref{sec:time}, that we pool together measurements and associate them with a single recovered frame.
This determines the capture duration per frame and its reciprocal, the temporal resolution.
The smaller the capture duration, the fewer the measurements that we acquire and hence, the greater the under-sampling.

These two metrics are linked by the measurement  rate of the camera and the resolution at which we operate the DMD.
For a capture duration of $\tau$ seconds, the under-sampling is given as 
$$ (N_1 N_2)/(\tau \times \textrm{measurement rate})$$
where the DMD resolution is $N_1 \times N_2$.
Recall, that the measurement rate of our LiSens prototype is nearly $1$ MHz while that of the SPC is $20$ kHz.

%
%Given that the DMD has a resolution of $1024 \times 768$, we multiplex along the axis with $768$ rows of micromirrors, we use $768$-dimensional column-permuted Hadamard matrix.
%
Finally, we map the $\pm 1$ entries of the Hadamard matrix to $0/1$  by mapping the `-1's to `0's;
in post-processing, the measurements corresponding to the $\pm 1$  matrix are obtained simply by subtracting the mean scene intensity.
For time-varying scenes, we repeat the all-ones measurement in the Hadamard matrix once every 100 measurements to track the mean intensity value.

\vspace{-4mm}
\paragraph{Gallery.} We present the images recovered by the LiSens camera on two indoor static scenes for various capture durations in Figure \ref{fig:gallery}. 
We also quantize the loss in performance with increased under-sampling using the reconstruction SNR.
Specifically, given a ground truth image $\bfx_{gr}$ and a reconstructed image $\widehat{\bfx}$, the reconstruction SNR in dB is given as 
$$ -20 \log_{10} \left( \frac{\| \bfx_{gr} - \widehat{\bfx}\|_2}{\| \bfx_{gr} \|_2}  \right).$$
We used the Nyquist rate reconstruction, i.e., under-sampling of $1\times$, as the ground truth image.
We observe that there is little loss in performance until a capture duration of $220 ms$ (under-sampling of $4\times$) and with a small drop in performance at $110 ms$ ($8\times$).
%
%This suggests a time resolution of $5-10$ frames per second.
%
As we will see later, using a spatio-temporal prior  improves the  reconstruction quality even at an under-sampling of $8 \times$.

\begin{figure*}[!ttt]
\includegraphics[width=\textwidth]{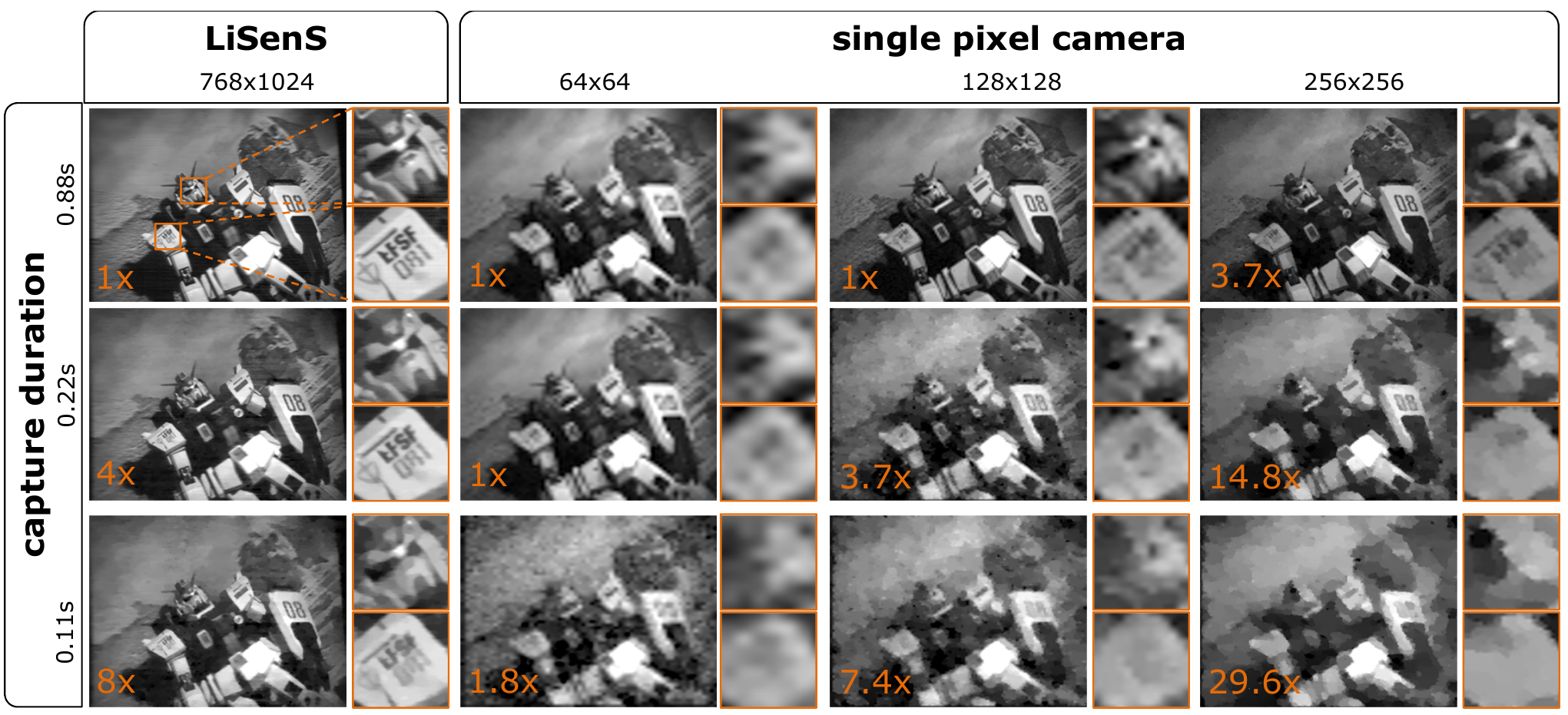}
\caption{\textbf{Performance of LiSens and SPC.}  We look at recovered images for various capture durations and, in the case of the SPC, various spatial resolution as well. The remarkable gap in the performance of the devices can be attributed to the difference in their measurement rates. Inset in orange are the under-sampling factors at which images were reconstructed.}
\label{fig:lisensvsspc}
\end{figure*}

\begin{figure*}[!ttt]
\center
\includegraphics[width=\textwidth]{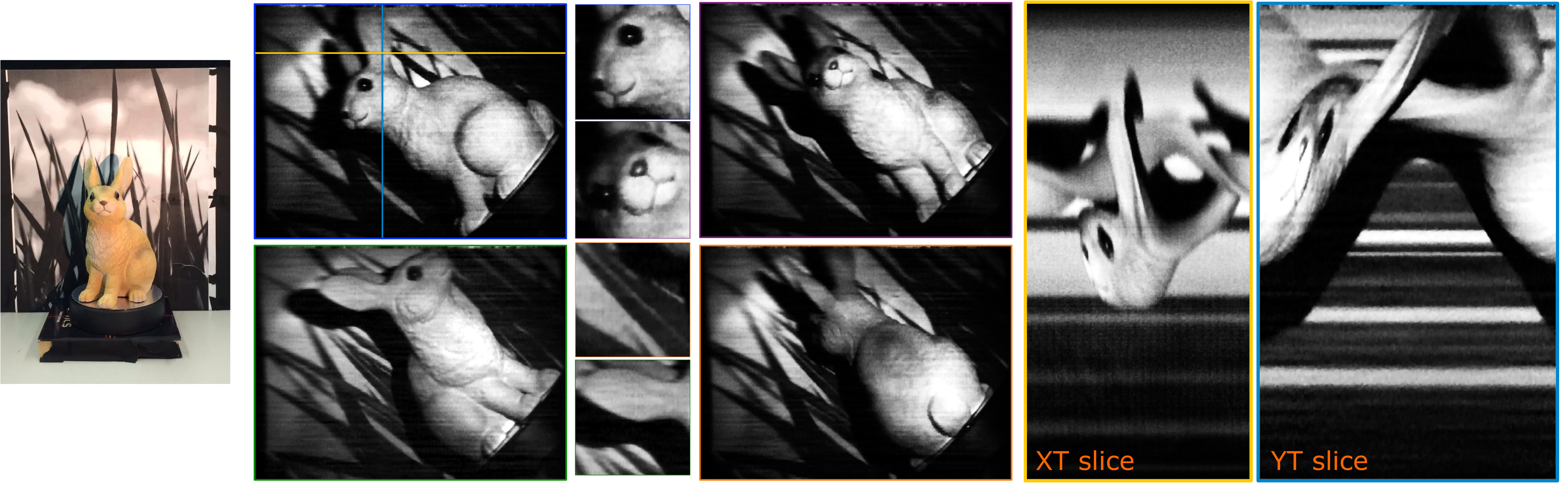}
\caption{\textbf{Video CS using LiSens}. We image a bunny on a turntable (left) and recover the video at a capture duration of $110 ms$ per frame (under-sampling of $8\times$) using a 3D TV prior. Shown are a few recovered frames and spatio-temporal slices. }
\label{fig:spatiotemporal}
\end{figure*}

\vspace{-4mm}
\paragraph{Comparison to SPC.} We compare the performance of LiSens and an SPC built on the off-axis of the DMD.
We image a static scene using both cameras and compare the reconstructed image for  varying capture durations (see Figure \ref{fig:lisensvsspc}). 
Recall that the measurement rate of the SPC is limited to 20 kHz while the LiSens has nearly 1 MHz.
This $50\times$ gap enables the LiSens camera to sense images at a resolution of $1024 \times 768$ at   $5-10$ frames per second.
As we will see next,  using simple spatio-temporal priors like a 3D TV norm will enable higher-quality reconstructions at an increased temporal resolution.
We note that the photodetector in the  SPC had a lower dynamic range than the line-sensor. For a fairer comparison, we took multiple runs with the SPC and averaged the measurements to artificially boost its dynamic range.

Figure \ref{fig:lisensvsspc} also indicates that LiSens operating at $8\times$ under-sampling  produces significantly higher quality results than the SPC operating at $7.4\times$ under-sampling. 
This is a consequence of the sparsity of images  scaling sub-linearly with image resolution. 
That is, if we looked at sparsity of images in a wavelet basis as a function of image resolution, the ratio of sparsity level to image resolution decreases monotonically. 
This can also be attributed to images having a lot of energy in low-freq components.
Hence, if we attempt to recover a scene at  $128\times128$  and $1024\times1024$, each at say $8\times$ under-sampling, then the quality would be significantly better for the high-resolution reconstruction.

\vspace{-4mm}
\paragraph{Dynamic scenes.} Next, we sensed a time-varying scene --- a bunny on a turntable spinning at  one revolution every 15 seconds.
%
%The scene was repeatable and hence, we could employ two distinct sensing strategies: Hadamard multiplexing in the absence of image priors, and compressive recovery using a 3D total variation prior.
%%
%First, for a given capture duration, we perform Hadamard multiplexing such that the under-sampling factor is exactly $1\times$; subsequently, we recover each frame individually by solving (\ref{eq:hada}).
%%
%Smaller capture times led to higher temporal resolution, albeit at a loss in spatial resolution.
%%
%This is seen in Figure \ref{fig:spatiotemporal}; for a capture time of $36 ms$ there is significant spatial blur while at $880ms$, there are significant artifacts due to motion blur and violation of the static scene assumption.
%%
%In between, at $110 ms$, we can obtain images at a resolution of $96 \times 1024$ and the two blur are equally present but are less perceptible; still there is a loss in spatial resolution as seen in the glint in the eye of the bunny.
%
%
We fix a capture duration of $110 ms$ per frame or equivalently a temporal-resolution of $9$ frames per second, and recover the frames of the video under a 3D total variation prior.
Due to high levels of measurement noise, the  recovered videos were subject to a $3\times 3 \times 3$ median-filter.
As seen in Figure \ref{fig:spatiotemporal}, the recovered video preserves fine spatial detail.
This can be attributed to the use of video priors, which capture inter-frame redundancies, as well as the high measurement rate enabled to our design.
This validates the potential of the LiSens architecture to acquire scenes at high spatial resolution ($1024 \times 768$) and with a reasonable temporal resolution.

\begin{figure*}[!ttt]
\center
\includegraphics[width=\textwidth]{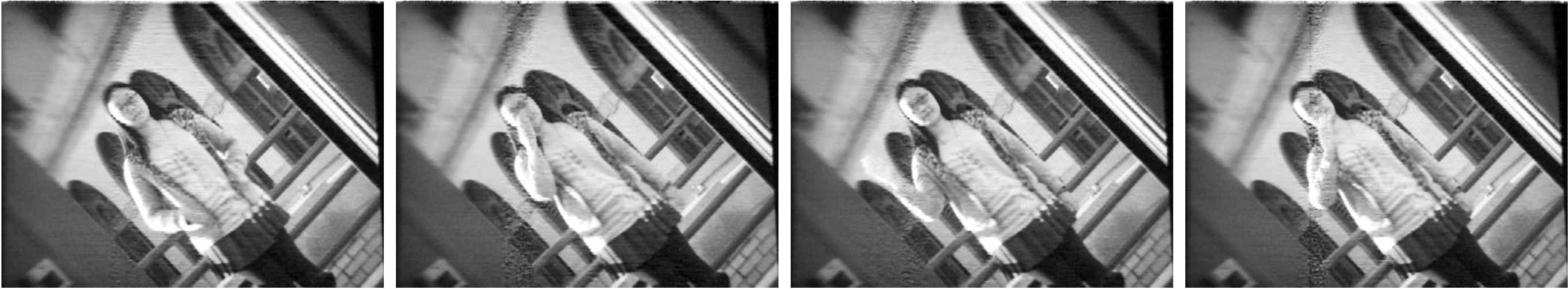}
\caption{\textbf{Sensing outdoors with the LiSens prototype.} }
\label{fig:outdoors}
\end{figure*}

Finally, Figure \ref{fig:outdoors} shows frames of a video of an outdoor scene reconstructed with the same parameters as  the indoor bunny scene.
The data was significantly noisier which contributed to the reduced reconstruction quality.
The use of stronger priors such as motion flow models as in \cite{reddy2011p2c2,sankaranarayanan2012cs} could  help in such scenarios.

\section{Discussion} \label{sec:discuss}
We present a multiplexing camera that is capable of delivering very high measurement rates, comparable to that of a full-frame sensor, but with a number of pixels that is a small fraction.
At its core, the proposed camera moves the bottleneck in the frame rate of the device from the spatial light modulator to the readout of the sensor.
The ensuing boost in measurement rate enables sensing of  scenes at video rate and at the full resolution of the light modulator.
Finally, while our prototype was built for visible wavebands, the underlying principles transfer mutatis mutandis to other wavebands of interest.

%\paragraph{Price.}  A SWIR line-detector from Hamamatsu (G10768-1024D) capable of a readout speed of 39,000 lines/second costs about USD 11k (this includes the readout circuit). With this architecture, we can obtain a measurement bandwidth of 39 MHz (39x greater than current prototype) and therefore potentially significantly better results with the caveat that faster-operation implies less light and hence lower SNR. The visible line-detector that we show in the paper costs about USD 2k. Other costs such as DMD {\it etc.} are significantly smaller if we do not buy development kits.

\section{Acknowledgement} \label{sec:Ack}
We thank Zhuo Hui  and Yang Gao for the help with the data collection.
J. W. and A. C. S. were supported in part by the NSF grant CCF-1117939.

\sloppy
{\small
\bibliographystyle{ieee}
\bibliography{videocs}
}
\end{document}